\documentclass{article}

\usepackage{microtype}
\usepackage{graphicx}
\usepackage{booktabs} 

\usepackage{multirow}
\usepackage{amssymb}
\usepackage{pifont}
\usepackage{amsmath}
\usepackage{threeparttable}
\usepackage{enumitem}
\usepackage{dirtytalk}
\usepackage{caption}
\usepackage{subfig}
\usepackage{todonotes}
\usepackage{stfloats}

\usepackage{hyperref}

\usepackage[accepted]{icml2020}

\usepackage{amsmath,amsfonts,amsthm,bm}

\def\1{\bm{1}}

\def\Secref#1{Section~\ref{#1}}

\def\eqref#1{equation~\ref{#1}}
\def\Eqref#1{Eq.~(\ref{#1})}

\def\Appendref#1{Appendix~\ref{#1}}

\DeclareMathAlphabet{\mathsfit}{\encodingdefault}{\sfdefault}{m}{sl}
\SetMathAlphabet{\mathsfit}{bold}{\encodingdefault}{\sfdefault}{bx}{n}

\def\0{{\bm{0}}}

\renewcommand{\P}{\mathbb{P}}
\newcommand{\E}{\mathbb{E}}

\newcommand{\N}{\mathcal{N}}

\newcommand{\Var}{\mathrm{Var}}

\icmltitlerunning{Making EfficientNet More Efficient}

\begin{document}

\twocolumn[
\icmltitle{Making EfficientNet More Efficient: Exploring Batch-Independent Normalization, Group Convolutions and Reduced Resolution Training}



\icmlsetsymbol{equal}{*}

\begin{icmlauthorlist}
\icmlauthor{Dominic Masters}{gc}
\icmlauthor{Antoine Labatie}{gcl}
\icmlauthor{Zach Eaton-Rosen}{gcl}
\icmlauthor{Carlo Luschi}{gc}
\end{icmlauthorlist}

\icmlaffiliation{gc}{Graphcore Research, Bristol, UK}
\icmlaffiliation{gcl}{Graphcore Research, London, UK}

\icmlcorrespondingauthor{Dominic Masters}{dominicm@graphcore.ai}

\icmlkeywords{Machine Learning, ICML}

\vskip 0.3in
]



\printAffiliationsAndNotice{}  

\begin{abstract}

Much recent research has been dedicated to improving the efficiency of training and inference for image classification. This effort has commonly focused on explicitly improving \textit{theoretical efficiency}, often measured as ImageNet validation accuracy per FLOP. These theoretical savings have, however, proven challenging to achieve in practice, particularly on high-performance training accelerators.

In this work, we focus on improving the \textit{practical efficiency} of the state-of-the-art EfficientNet models on a new class of accelerator, the Graphcore IPU. We do this by extending this family of models in the following ways: (i) generalising depthwise convolutions to group convolutions; (ii) adding proxy-normalized activations to match batch normalization performance with batch-independent statistics; (iii) reducing compute by lowering the training resolution and inexpensively fine-tuning at higher resolution. We find that these three methods improve the practical efficiency for both training and inference. Code available at \url{https://github.com/graphcore/graphcore-research/tree/main/Making_EfficientNet_More_Efficient}.
\end{abstract}

\section{Introduction}
\label{intro}

Making computer vision models more \textit{efficient} would allow them to exceed the performance of current models given any specified constraints, including: final accuracy, energy consumption, time-to-train, inference latency, model size, total cost of training and more. One commonly-used proxy for the \textit{efficiency} is the total number of floating-point operations (FLOPs) used. However, there is often a significant disparity between the efficiency one would expect from the FLOP count (which we describe as the \textit{theoretical efficiency}) and the observed performance on modern hardware accelerators. This disparity holds especially for training \cite{Touvron21, Lee2020}. One reason for this is that not all FLOPs are created equal: optimising the FLOP count can often have the side effect of requiring additional data movement or poorly utilising vectorised hardware instructions.
This can adversely affect the total FLOP rate. In this work, we aim to improve the \textit{practical efficiency} of EfficientNet B0-B5, which were originally optimised for FLOPs, and better realise their promise as low-resource, high-performing networks. 
To this end, we consider three algorithmic improvements to EfficientNet:
 
\begin{figure}[tb]
\centering
\includegraphics[width=1\linewidth]{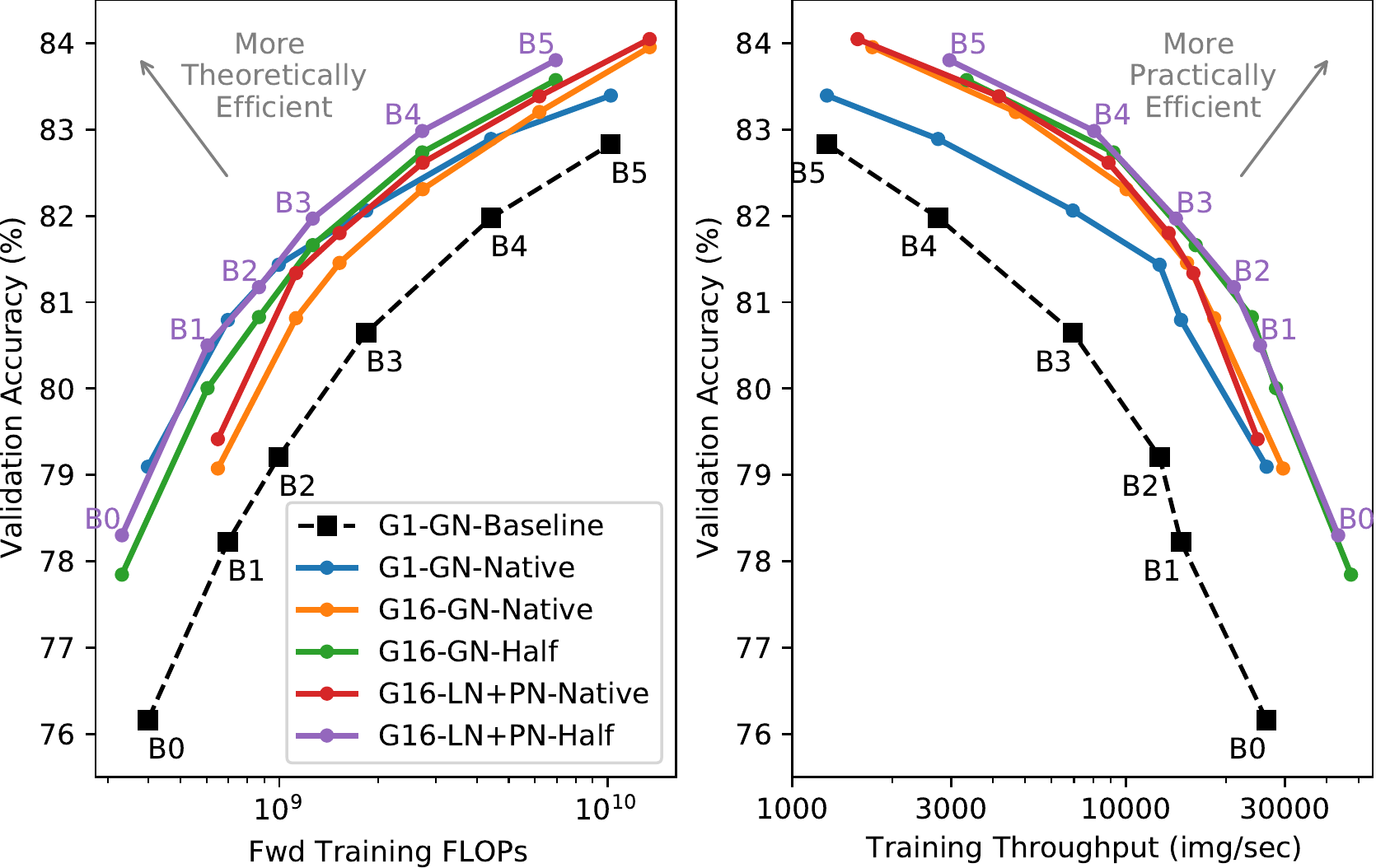}\\
\caption{Theoretical (left) and practical (right) training efficiency of EfficientNet B0-B5 models using our three proposed improvements: (i) group convolutions [G16 (ours) vs G1]; (ii) proxy-normalized activations [LN+PN (ours) vs GN] and (iii) half-resolution training  [Half (ours) vs Native]. Results show the best accuracy achieved when fine-tuning and testing on a range of resolutions. Baseline has no fine-tuning and uses the \textit{native} image resolution.}
\label{fig:front_page}
\end{figure}

 \begin{itemize} 
 \setlength\itemsep{0em}
 \item Generalising depthwise convolutions to group convolutions; 
 \item Applying \textit{proxy-normalized activations} to bridge the performance gap between batch-independent normalization methods and Batch Normalization;
 \item Reducing compute by lowering the training resolution and inexpensively fine-tuning at higher resolution.
 \end{itemize} 
 We find that all three methods improve training accuracy and/or throughput such that their combination increases training and inference throughput by up to $7\times$ and $3.6\times$, respectively, and reduces inference latency by up to $1.4\times$. 

\section{Background}
\label{background}
\subsection{Efficient CNNs}
\label{sec:eff-cnns}

Practical training efficiency improvements have been a significant enabler of innovation throughout the evolution of CNNs. For example, the innovation of AlexNet~\cite{Krizhevsky12} was largely enabled due to GPU acceleration, and arguably, the success of ResNet~\cite{He15} can be attributed not just to its good task performance but also due to its high throughput on GPUs relative to alternative models. 

More recently, major improvements in terms of theoretical efficiency have been achieved. The most notable innovation has been the introduction of group and depthwise convolutions for spatial operations. Introducing group convolutions alone was found to improve the theoretical efficiency of ResNet-50~\cite{Xie16, Ioannou16}. Similarly, by reducing the group size to 1, \citet{Chollet16} leveraged depthwise convolutions to achieve gains in theoretical efficiency with respect to analogous models with dense convolutions. In particular, this led to major advances in compute- and memory-constrained \say{mobile} applications~\cite{Howard17,Zhang17,Iandola16,Sandler18}. 

These theoretical efficiency benefits were further enhanced by directly minimising FLOPs using Neural Architecture Search (NAS). This led to efficiency improvements across the full spectrum of model sizes, from mobile-sized models like MobileNetV3~\cite{Howard19} and MNasNet~\cite{Tan18} to large models like NASNet~\cite{Zoph17} and AmoebaNet \cite{Real18}.  
Notably, all NAS models that feature in the top-100 highest accuracy ImageNet models use group or depthwise convolutions in some form, further highlighting the advantage of these operations over their dense counterparts.\footnote{\url{https://sotabench.com/benchmarks} accessed 4th February 2021}
Building on the already efficient MNasNet, EfficientNet~\cite{Tan19} added improved training methods and scaling to much larger models to achieve SOTA performance across a broad spectrum of FLOP budgets. 

While practical improvements were often achieved with \say{efficient} models for low power CPUs, these models have typically struggled to convert the theoretical gains into higher training throughput on high-performance hardware~\cite{Touvron21, Lee2020}. For example, while EfficientNets vastly outperform ResNets in terms of theoretical training efficiency, they have often been found to underperform when considering practical training efficiency on GPUs~\cite{Lee2020}. Some recent work has used NAS to optimise practical efficiency on GPUs~\cite{Cai18, Vahdat19, Lin20}. For the presented work, we prioritised hang-engineered solutions, but do not rule out NAS methods in future work. 

\subsection{Hardware Considerations}
\label{sec:hardware}
When investigating the practical efficiency of a model, it is important to understand the characteristics of the hardware it runs on. 
Discussion around this often focuses heavily on the peak compute rate, measured in floating-point operations per second (FLOPS), the theoretical maximum rate of compute operations. While the peak rate is an important factor to consider, it can be equally important to understand the assumptions required to achieve it -- for example, the structure of the compute and the availability of data.

The compute structure is important as modern hardware typically utilises vector instructions that allow dot products of a given length to be computed with a single instruction. If, however, the compute cannot be structured such that these vector instructions are filled, FLOPs can potentially be wasted. Furthermore, if the data is not immediately available at the compute engine, then cycles will be required to move it. The extent to which this limitation manifests will be highly dependent on the \textit{memory bandwith}.

The reliance on memory bandwidth depends on the model and can be characterised by the ratio of compute to data transfer, i.e., the \textit{arithmetic intensity} -- where low arithmetic intensity operations are much more dependent on memory bandwidth.
For a simple group convolution, arithmetic intensity monotonically increases with group size, kernel size, field size and batch size (see Appendix \ref{sec:arithemtic_intensity}). Notably, this means that the efficiency of depthwise convolutions and group convolutions with small group size is more likely to be limited by the available memory bandwidth.

In this work, we use a new class of hardware accelerator, the Graphcore IPU. This accelerator has many characteristics that set it aside from the GPUs that are commonly used for neural network training. The IPU compute is distributed across the chip in 1472 cores and, while its instructions are still vectorised, only dot products of 16 terms are required to utilise the compute engines fully. This helps to reduce the dependence on the compute structure. Furthermore, the IPU has over 900MB of high-bandwidth on-chip memory, significantly more than alternative hardware. This sharply reduces the cost of low arithmetic intensity operations. 

To maximise performance on the IPU, it becomes important to keep as much of the working memory -- for example, activation state -- on-chip. This naturally promotes the use of much smaller batches, memory saving optimisation \cite{Chen16a,Gruslys16}, and innovative forms of distributed processing \cite{Harlap18,Huang19,Ben-Nun18,Shazeer18}. At the same time, it does require reconsidering the use of Batch Normalization~\cite{Ioffe15}, the most common normalization method in vision models, which relies on large batches.

\section{Methods}
\label{sec:Methods}

\subsection{Group Convolutions}
\label{sec:methods-group-convs}

As stated in \Secref{sec:eff-cnns}, NAS methods tend to \textit{group} their spatial convolutions, often with group size $G=1$ (depthwise convolutions). While depthwise convolutions are very FLOP and parameter efficient~\cite{Howard17}, using $G>1$ would utilise modern hardware accelerators more efficiently as a larger group size: (i) increases the arithmetic intensity (see Appendix~\ref{sec:arithemtic_intensity}); (ii) increases the length of the dot products (used in the convolutions), allowing larger vector instructions to be utilised. 

We aim to investigate the trade-offs involved in increasing the group size $G$ of the spatial convolutions for the EfficientNet model. Increasing $G$ alone would increase both parameter count and FLOPs. Therefore to maintain similar model complexity we correspondingly decreasing the \textit{expansion ratio}, defined as the ratio of channels between the input to the first pointwise convolution and the spatial convolution.
This is similar to the FLOP-equivalent scaling of the ResNeXt family~\cite{Xie16}. Therefore, a network with larger $G$ will be \textit{narrower} for the same FLOP count, yielding computational benefits by reducing the size of the stored activation state and facilitating the use of larger batch sizes. Note that while this compensation aims to keep the total FLOPs and parameters count similar to the baseline, for simplicity, we change the expansion ratio only at a global level. Consequently, we do not necessarily maintain the exact same distribution of parameters and FLOPs with depth.

As in EfficientNet, other NAS-derived architectures typically only use depthwise convolutions, which suggests that depthwise convolutions were optimal with respect to validation accuracy. In ResNeXts, increasing $G$ while maintaining total FLOPs leads to decreasing validation accuracy. This would also suggest that vanilla EfficientNet, with $G=1$, would achieve higher accuracy compared to similar networks with $G>1$. Nevertheless, we hope that the modified networks offer better trade-offs between task performance and training time. We, therefore, test group sizes between $G=1$ and $G=64$ for EfficientNet B0 and B2.\footnote{For layers where the group size does not divide the channels, the group size is rounded to the nearest value that does.} 

\subsection{Batch-Independent Normalization}
\label{sec:norm}
As mentioned in Section \ref{sec:hardware}, vision models commonly rely on the use of Batch Normalization (BN)~\cite{Ioffe15} to normalize activations throughout the network. BN is typically applied on the \textit{unnormalized pre-activations} $X$ to produce the \textit{normalized pre-activations} $Y$, before an affine transform and a nonlinearity $\phi$ finally produce the \textit{post-activations} $Z$. Formally, for each channel $c$:
\begin{align}
    Y_{\cdot\cdot\cdot c} & = \frac{X_{\cdot\cdot\cdot c} -\mu_c}{\sqrt{\sigma^2_c+\epsilon}}, \label{eq:bn} \\  
    Z_{\cdot\cdot\cdot c} & =\phi\big(\gamma_c Y_{\cdot\cdot\cdot c}+\beta_c\big), \label{eq:act}
\end{align}
where $\cdot$ is an index placeholder, $\epsilon$ is BN's numerical stability constant, $\mu_c$, $\sigma_c$ are the mean and standard deviation of $X$ in channel $c$, and $\gamma_c$, $\beta_c$ are BN's scale and shift parameters restoring in each channel $c$ the two degrees of freedom lost in \Eqref{eq:bn}.

The normalization of BN in \Eqref{eq:bn} ensures that $Y$ is normalized, by which we mean that it has zero mean and unit variance in each channel $c$. This foundational principle of BN is essential for the successful scaling to large and deep models:
\begin{enumerate}[label=(\roman*)]
\setlength\itemsep{0em}
\item By ensuring that the nonlinearity $\phi$ \say{sees} a data distribution close to normalized in each channel,\footnote{We say that a distribution is \say{close to normalized} when its mean is not far from zero and its variance not far from one. Since $\gamma_c Y_{\cdot\cdot\cdot c}+\beta_c$ and the normalized $Y_{\cdot\cdot\cdot c}$ are only separated by an affine transform with parameters not subject to weight decay, we implicitly assume that $\gamma_c Y_{\cdot\cdot\cdot c}+\beta_c$ is \say{close to normalized}.} $\phi$ can effectively be nonlinear \emph{with respect to this distribution}. Consequently, additional layers can add expressive power and the network can effectively use its whole depth. This is opposed to a situation where $\phi$ would \say{see} a \say{collapsed} data distribution, such that it would become well approximated at first order by a linear function with respect to this distribution;
\item By ensuring that different channels have close to equal variance, the network can effectively use its whole width. This is opposite to a situation where a single channel would become arbitrarily dominant over the others, such that it would become the only channel \say{seen} by subsequent layers.
\end{enumerate}

Despite the practical success deriving from this foundational principle, the reliance of BN on the mini-batch of data can sometimes be problematic. Most notably, when the mini-batch is small or when the dataset is large, the regularisation coming from the noise in the mini-batch statistics $\mu_c$, $\sigma_c$ can be excessive or unwanted, leading to degraded performance \citep{Ioffe17,Wu18,Masters18,Yin18,Luo18,Kolesnikov20,Summers20}.

To circumvent these problems, a variety of batch-independent normalization techniques have been proposed in the literature: Layer Normalization (LN)~\cite{Ba16}, Group Normalization (GN)~\cite{Wu18}, Instance Normalization (IN)~\cite{Ulyanov16}, Weight Normalization (WN)~\cite{Salimans16}, Weight Standardization (WS)~\cite{Qiao19}, Online Normalization (ON)~\cite{Chiley19}, Filter Response Normalization (FRN)~\cite{Singh19}, EvoNorm~\cite{Liu20}, being a non-exhaustive list. 
While useful in other contexts, none of these techniques managed to close the performance gap with large-batch BN in the context of this work, focused on EfficientNets trained with RMSProp on ImageNet. 

This led us to rethink how to perform batch-independent normalization and, in concurrent work, propose \textit{Proxy-Normalized Activations}~\cite{Labatie21}. 
In this work \citet{Labatie21} formulate the postulate that, beyond lifting the reliance on the micro-batch, batch-independent normalization should also maintain BN's principle of normalizing pre-activations $Y$ in each channel. A first justification of this postulate is drawn from the benefits that are expected -- as explained above -- from this inductive bias of BN. A second justification is drawn, on a more practical level, from the fact that BN was used in architecture searches such as the one that yielded the EfficientNet family of models. Sticking to the same normalization principles might therefore spare the need of redoing these searches.

To retain BN's principles while removing any dependence on the batch size, \citet{Labatie21} extend the work of \citet{Arpit16} as follows: 
(i) replacing the BN step of \Eqref{eq:bn} by a batch-independent normalization step, based on either LN or GN; 
(ii) replacing the activation step of \Eqref{eq:act} by a \textit{proxy-normalized activation} step. This step normalizes $\phi(\gamma_c Y_{\cdot\cdot\cdot c} + \beta_c)$ by assimilating it with $\phi(\gamma_c \tilde{Y}_c + \beta_c)$, where $\tilde{Y}_c\sim\N\big(\tilde{\beta}_c,(1+\tilde{\gamma}_c)^2\big)$ is a Gaussian \textit{proxy} variable with mean $\tilde{\beta}_c$ and variance $(1+\tilde{\gamma}_c)^2$ derived from the additional parameters $\tilde{\beta}_c$, $\tilde{\gamma}_c$. $\tilde{\beta}_c$, $\tilde{\gamma}_c$ are subject to weight decay to express the prior that $Y$ is close to normalized. If LN is chosen as the batch-independent normalization, this is expressed for each batch element $b$ and channel $c$ as:
\begin{gather}
    Y_{b\cdot\cdot\cdot} = \frac{X_{b\cdot\cdot\cdot} -\mu_b}{\sqrt{\sigma^2_b+\epsilon}}, \label{eq:ln} \\
    Z_{\cdot\cdot\cdot c} = \frac{\phi\big(\gamma_c Y_{\cdot\cdot\cdot c}+ \beta_c\big) - \E_{\tilde{Y}_c}\big[\phi\big(\gamma_c \tilde{Y}_c  +\beta_c\big)\big]}{\sqrt{\Var_{\tilde{Y}_c}\big[\phi\big(\gamma_c \tilde{Y}_c +\beta_c\big)\big]+\tilde{\epsilon}}}, \label{eq:pn}
\end{gather}

where $\tilde{Y}_c \sim \N\big(\tilde{\beta}_c,(1+\tilde{\gamma}_c)^2\big)$, $\epsilon$, $\tilde{\epsilon}$ are respectively LN's and proxy normalization's numerical stability constants, and $\mu_b$, $\sigma_b$ are the mean and standard deviation over the spatial and channel dimensions of the batch element $b$ of $X$.

When combined with LN, such Proxy Normalization (PN) of activations iteratively ensures that pre-activations $Y$ remain close to normalized. This is corroborated both theoretically and experimentally in \citet{Labatie21}.

\subsection{Image Resolution}
\label{sec:methods-image-res}

The introduction of global average pooling~\cite{Lin13} allowed image classification CNNs to operate on inputs of arbitrary resolution. While this has been explored in tasks such as image segmentation~\cite{Long14}, in image classification, its impact is still to be fully understood. The EfficientNet model treats the image resolution as a tuneable hyperparameter, using larger images to train larger networks. \citet{Hoffer19} trained networks on several images sizes simultaneously, leading to either i) accelerated training to a target accuracy or ii) improved final performance for the same training resources. Perhaps closest to our aims, \citet{howard_training_imagenet} recommends starting training with low-resolution images, increasing their size progressively during training, with the goal of reducing the total time-to-train. 

In recent works, \citet{Touvron19, Touvron20} showed that small amounts of fine-tuning can enable a network to process test images at higher resolution than during training.  
The fine-tuning step only needs to act on the final portion of the network, and only for a few epochs, to increase the overall accuracy. 
Consequently, the computational cost of fine-tuning is almost negligible in comparison to the rest of training. 

We take inspiration from this to investigate fine-tuning of a network trained on low-resolution images and generalising it to larger resolutions for the perspective of efficiency. The use of smaller images during training allows us to train a given model faster, using less memory, or to train a larger model in the same amount of time. 
To test this idea, we compare training at the native EfficientNet image size (as defined in~\citet{Tan19} -- reproduced in Table~\ref{tab:resolutions}) to training with approximately half the original number of pixels (width and height therefore approximately $1/\sqrt{2}$ times the original), which we denote as \textit{half resolution}.
In particular, this approximately matches the FLOPs of the EfficientNet model one size down.

We then fine-tune and test at a range of image sizes up to $700\times700$. 
When choosing the precise resolution to use for validation, we note that performance can suffer from an aliasing artefact. This artefact arises due to the locations of asymmetric downsampling layers, where the input field dimensions are odd, which happens at different depths depending on the input resolutions. We find that it is important to keep the location of these downsampling layers consistent between training and testing. This can be achieved by choosing the test resolution, $r_{test}$, such that
$r_{train} \equiv r_{test} \! \pmod{2^n}$
where $n$ is the number of downsampling layers in the model ($n=5$ for EfficientNet). 

We also choose our \textit{half resolution} to approximately halve the total number of pixels from the original EfficientNet training resolutions~\citep{Tan19}, while satisfying this condition (see Table~\ref{tab:resolutions}). This allows for direct comparisons to be made between the two training regimes.

\begin{table}[htb]
\caption{Image sizes used for training in native and half resolution.}
\label{tab:resolutions}
\vskip 0.15in
\begin{center}\begin{small}\begin{sc}
\scalebox{1.}{
\begin{tabular}{lcccccc}
\toprule
       & B0  & B1  & B2  & B3  & B4  & B5  \\
\midrule
Native & 224 & 240 & 260 & 300 & 380 & 456 \\
Half   & 160 & 176 & 192 & 204 & 252 & 328 \\
\bottomrule
\end{tabular} 
}
\end{sc}\end{small}\end{center}
\vskip -0.15in
\end{table}

\subsection{Implementation Details}
\label{sec:implementation}

Throughout this work, unless otherwise stated, we run our experiments on a Graphcore POD16 server. This contains 4$\times$ M2000s, that have each 4$\times$ MK2 IPU chips along with 2$\times$ AMD EPYC 7742 64-core CPUs. Each IPU chip has 1472 cores, with 642KB of local SRAM for each core, totalling in excess of 900MB of high-bandwidth on-chip memory. Further storage is provided by off-chip DDR4. The results are generated using the TensorFlow\cite{tensorflow2015-whitepaper} framework with Poplar SDK v2.0\footnote{\url{https://www.graphcore.ai/products/poplar}} 

During training, the model is divided into several \textit{stages} that are processed in a pipeline across 2 or 4 chips, to maximise IPU utilisation. Following the pipelined paradigm, we do not process an entire mini-batch simultaneously. Instead, we sequentially compute and accumulate gradients for many smaller \textit{micro-batches}, before applying the weight update with global batch size $B$. Our model-parallel pipelined implementation reduces the memory requirements on each chip, allowing the use of larger micro-batches and thus increasing throughput. This scheme is similar to~\citet{Huang19}, but differs from~\citet{Harlap18} in the detail that we recompute each forward stage before its respective backwards pass. We replicate this setup to use all available IPUs in a data-parallel configuration, allowing us to use all 16 IPUs. The arithmetic is computed in float-16 mixed-precision, with each weight's optimiser state stored in float-32~\cite{Micikevicius17}. For inference, a single IPU is used with all arithmetic operations and weights in float-16. 

Our baseline EfficientNet model architecture is as described in \citet{Tan19}, except for the replacement of BN layers with GN with 4 groups~\cite{Wu18}. Thus, our training dynamics depend on the global mini-batch size but not the micro-batch size (unlike if we had used BN). 
We find that good task performance depends on the degree of regularisation we use. While the original EfficientNet paper uses AutoAugment~\cite{Cubuk18}, we find that the augmentation hyperparameters from the original work are not robust to small model modifications (see Table~\ref{tab:tpu}), which agrees with conclusions made in~\citet{Brock21}. Furthermore, AutoAugment is computationally costly and can make CPU performance the throughput bottleneck.
We, therefore, follow the augmentation strategy of~\citet{Brock21} by using a combination of Mixup~\cite{Zhang17a} and CutMix~\cite{Yun19}. We find it beneficial to increase the strength of these methods as model size increases. Otherwise preprocessing steps follow the procedure of \citet{He15}.

The training procedure closely follows~\citet{Tan19}. We train on ImageNet~\cite{Russakovsky15} for 350 epochs with RMSProp~\citep{Tieleman12} and decay the learning rate exponentially by a factor 0.97 every 2.4 epochs. We use a weight decay of $10^{-5}$ on the convolutional weights, a label smoothing factor of $0.1$, and normalization and PN's numerical stability constants $\epsilon=10^{-3}$ and $\tilde{\epsilon}=3\times10^{-2}$.
We use a slightly smaller global batch size $B=768$ across all training cases and scale the original learning rate and RMSProp decay factor. For the RMSprop optimiser we use learning rate $B\times2^{-14}$,  momentum $0.9$ and decay $1.0-B\times 2^{-14}$. 
Our final weights are obtained by using an exponentially weighted average over checkpoints from each training epoch, with decay factor $0.97$. 

For fine-tuning we train, starting from the weight averaged checkpoint, for an additional 2 epochs, using vanilla Stochastic Gradient Descent (SGD) with a global batch size $B=512$. We use a cosine learning rate schedule with an initial learning rate of 0.25 and the same preprocessing as for validation, as suggested in \citet{Touvron19}. We consider three different depths to fine-tune from: (i) the \textit{last block} which contains the final convolution, normalization and fully connected layer weights; (ii) the \textit{last two blocks} which also include all weights up to and including the previous downsampling layer; or (iii), the \textit{last three blocks} which contain all weights back one further downsampling layer. This was done to keep the \textit{proportion} of layers that are fine-tuned approximately independent of the model size. 

All our accuracy results derive from an averaging over three independent runs. Real data is used for all training throughput measurements. For inference cases, we use synthetic data as a real production inference pipeline would differ from the setup we use in this work, which primarily aims to calculate validation accuracy.

\section{Results}
\label{sec:results}
\subsection{Group Convolutions}
\label{sec:results-gc}

In Table~\ref{tab:group_conv}, we present the results of our experiments for different group and network sizes. We use EfficientNet B0 and B2 as a test-bed and sweep a range of group sizes. The experiments show that the validation performance for $G\in\{4, 16\}$ matches or exceeds the vanilla baseline performance.

While the group size $G=4$ case achieves the best overall validation accuracy in these tests, we find that the increased computational benefits of group size $G=16$ yields a superior trade-off in practice. We refer to this model as \textit{G16-EfficientNet}. We leave further investigation of the \textit{G4-EfficientNet} to future work.

Testing the \textit{G16-EfficientNet} for EfficientNet B1, B3, B4 and B5 (Table~\ref{tab:group_conv}), we find that the validation accuracy is actually higher for $G=16$ than for the $G=1$ baseline. This result exceeds our expectations and suggests that this variant is not only faster but also more accurate.
When testing on the full range of image sizes, we see further performance improvements for \textit{G16-EfficientNet} over the group size $G=1$ baseline (Table~\ref{tab:Train_Test_results_N}-\ref{tab:Train_Test_results_B}, Figure~\ref{fig:Train_Test_a}). 

\begin{table}[htb]
\begin{center}
\caption{Validation accuracy $\pm$ standard deviation (\%) for different group sizes. All training and validation experiments performed at the native image size, with Group Normalization and without fine-tuning.}
\label{tab:group_conv}
\vskip 0.15in
\begin{threeparttable}
\begin{small}
\begin{sc}
\begin{tabular}{lllllr}
\toprule
Size &   G &  E &     P &     F &     Accuracy \\
\midrule
\multirow{5}{*}{B0} &   1 &  6 &   5.3 &   0.4 &  76.2{\tiny$\pm$0.1} \\
 &   4 &  5 &   5.1 &   0.4 &  76.4{\tiny$\pm$0.1} \\
 &  16 &  4 &   5.9 &   0.6 &  76.2{\tiny$\pm$0.3} \\
 &  32 &  3 &   6.2 &   0.9 &  75.4{\tiny$\pm$0.1} \\
 &  64 &  2 &   6.7 &   1.5 &  73.9{\tiny$\pm$0.1} \\
\midrule
\multirow{2}{*}{B1} &   1 &  6 &   7.8 &   0.7 &  78.2{\tiny$\pm$0.1} \\
 &  16 &  4 &   8.3 &   1.1 &  78.3{\tiny$\pm$0.1} \\
\midrule
\multirow{5}{*}{B2} &   1 &  6 &   9.1 &   1.0 &  79.2{\tiny$\pm$0.2} \\
 &   4 &  5 &   8.6 &   1.0 &  79.7{\tiny$\pm$0.1} \\
 &  16 &  4 &   9.5 &   1.5 &  79.4{\tiny$\pm$0.1} \\
 &  32 &  3 &  10.3 &   2.1 &  78.5{\tiny$\pm$0.2} \\
 &  64 &  2 &   9.9 &   3.6 &  76.9{\tiny$\pm$0.1} \\
\midrule
\multirow{2}{*}{B3} &   1 &  6 &  12.2 &   1.8 &  80.6{\tiny$\pm$0.0} \\
 &  16 &  4 &  12.6 &   2.7 &  80.9{\tiny$\pm$0.1} \\
\midrule
\multirow{2}{*}{B4} &   1 &  6 &  19.3 &   4.4 &  82.0{\tiny$\pm$0.0} \\
 &  16 &  4 &  19.3 &   6.2 &  82.3{\tiny$\pm$0.1} \\
\midrule
\multirow{2}{*}{B5} &   1 &  6 &  30.4 &  10.2 &  82.8{\tiny$\pm$0.1} \\
 &  16 &  4 &  28.7 &  13.4 &  83.4{\tiny$\pm$0.1} \\
\bottomrule
\end{tabular}
\end{sc}
\begin{tablenotes}
\item[*] G = group size, E = expansion ratio, \\P = parameters (millions), F = FLOPs (billions)
\end{tablenotes}
\end{small}
\end{threeparttable}
\end{center}
 \vskip -0.1in
\end{table}

\begin{table*}[tb]
\caption{Comparison of validation accuracy $\pm$ standard deviation (\%) with Batch Normalization (BN), Group Normalization (GN) and Layer Normalization + Proxy-Normalized Activations (LN+PN). BN results show validation accuracy when training with baseline preprocessing and AutoAugment, respectively. BN results generally reproduced from~\cite{Tan18}, with our BN results marked by * (see \Appendref{sec:TPU Experiments})}
\label{tab:norm}
\vskip 0.15in
\begin{center}
\begin{small}
\begin{sc}
\scalebox{1}{
\begin{tabular}{lcccccc}
\toprule
{} &      \multicolumn{2}{c}{$G=1$, $E=6$} & \multicolumn{4}{c}{$G=16$, $E=4$} \\
\cmidrule(lr){2-3} \cmidrule(lr){4-7}
{} &         BN &     BN (AA) & BN &     BN (AA) &    GN & LN+PN \\
\midrule
B0   &  76.9{\tiny$\pm$0.1}* & 77.2{\tiny$\pm$0.1}* & 76.8{\tiny$\pm$0.1}* & 76.7{\tiny$\pm$0.1}* &  76.2{\tiny$\pm$0.3} &  76.8{\tiny$\pm$0.1} \\
B1   &  78.7 & 79.1 &   -   & - &  78.3{\tiny$\pm$0.1} &  79.0{\tiny$\pm$0.1} \\
B2   &  79.4{\tiny$\pm$0.0}* & 80.0{\tiny$\pm$0.0}* & 79.5{\tiny$\pm$0.1}* & 79.7{\tiny$\pm$0.1}* &  79.4{\tiny$\pm$0.1} &  79.9{\tiny$\pm$0.1} \\
B3   &  81.1 & 81.6 &   -   & - &  80.9{\tiny$\pm$0.1} &  81.3{\tiny$\pm$0.1} \\
B4   &  82.5 & 82.9 &   -   & - &  82.3{\tiny$\pm$0.1} &  82.6{\tiny$\pm$0.1} \\
B5   &  83.1 & 83.6 &   -   & - &  83.4{\tiny$\pm$0.1} &  83.4{\tiny$\pm$0.1} \\
\bottomrule
\end{tabular}
}
\end{sc}
\end{small}
\end{center}
\end{table*}

\begin{figure}[tb]
\centering
\includegraphics[width=.9\linewidth]{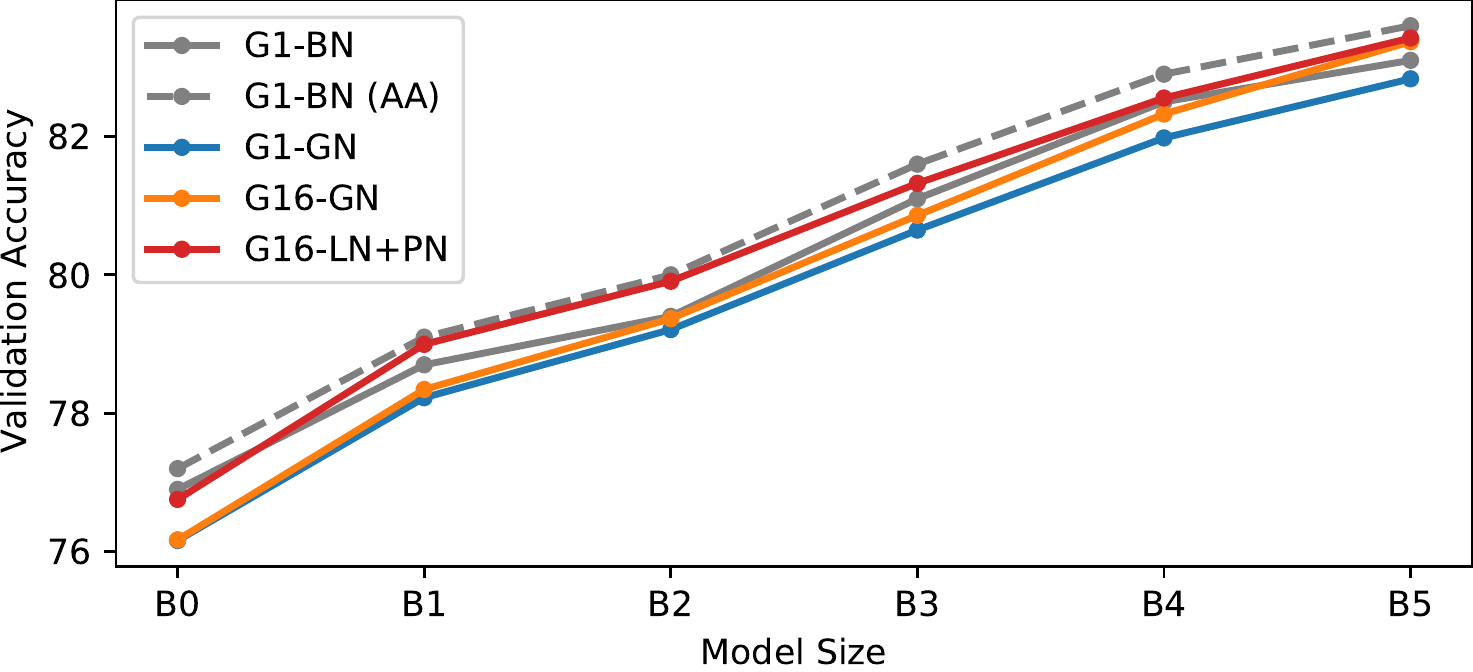}\\
\caption{Validation accuracy with Group Normalization (GN) and Layer Normalization + Proxy-Normalized Activations (LN+PN) compared with Batch Normalization (BN) for different model sizes.}
\label{fig:Norm}
\end{figure}

\subsection{Proxy-Normalized Activations}
\label{sec:pn}
Table \ref{tab:norm} and Figure \ref{fig:Norm} show that accuracy results obtained with LN+PN match accuracy results obtained with BN across all model sizes. This conclusion holds if we assume that: (i) our regularisation strategy is an \say{intermediate} between the baseline preprocessing and the computationally expensive AutoAugment; (ii) accuracy results with BN are similar in the $G=1$ and $G=16$ cases. In any case, Table \ref{tab:norm} shows that the accuracy achieved with LN+PN matches the best accuracy obtained with BN when directly comparing the two methods on $G=16$ for both B0 and B2. 

Beyond theoretical considerations, the choice of combining LN with PN is thus backed by experimental results.

\subsection{Train-Test Image Resolution}
\label{sec:train_test}

\begin{figure*}[ht]%
\centering
\subfloat[a][]{\label{fig:Train_Test_a}\includegraphics[width=0.31\linewidth]{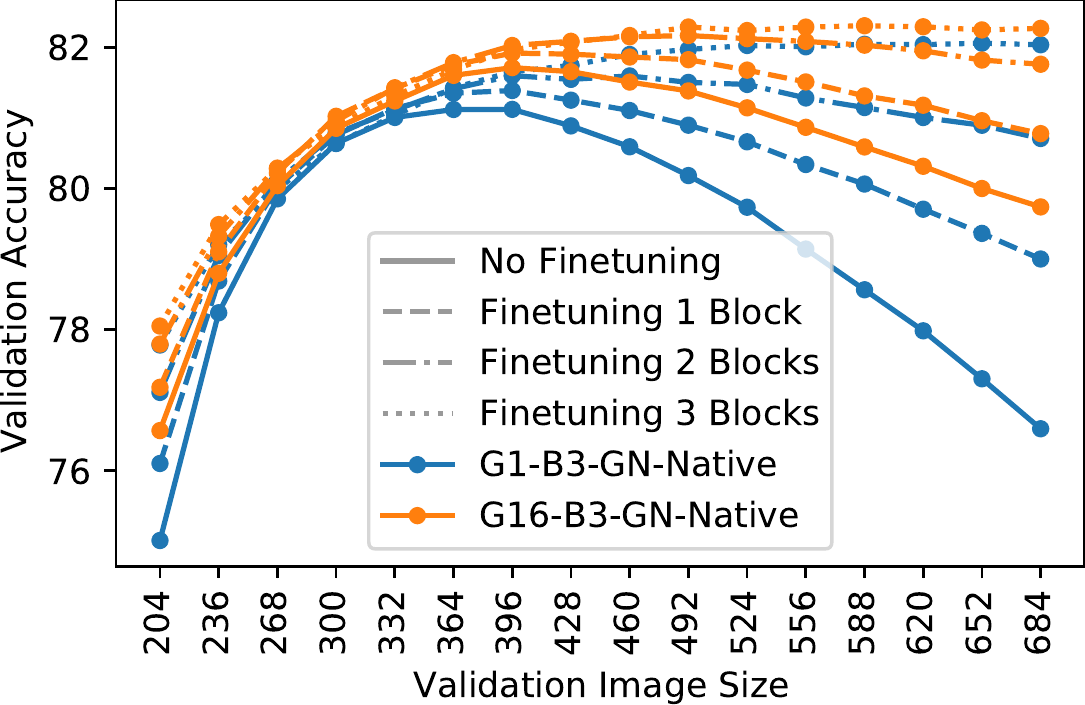}}\quad
\subfloat[b][]{\label{fig:Train_Test_b}\includegraphics[width=0.31\linewidth]{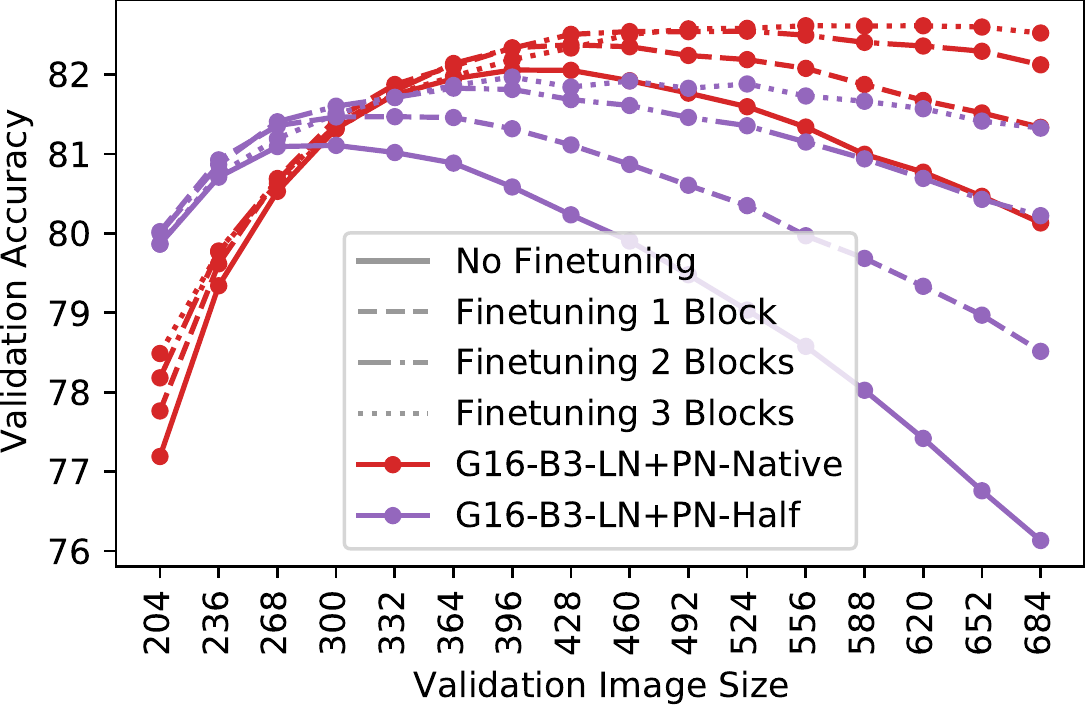}}\quad
\subfloat[c][]{\label{fig:Train_Test_c}\includegraphics[width=0.31\linewidth]{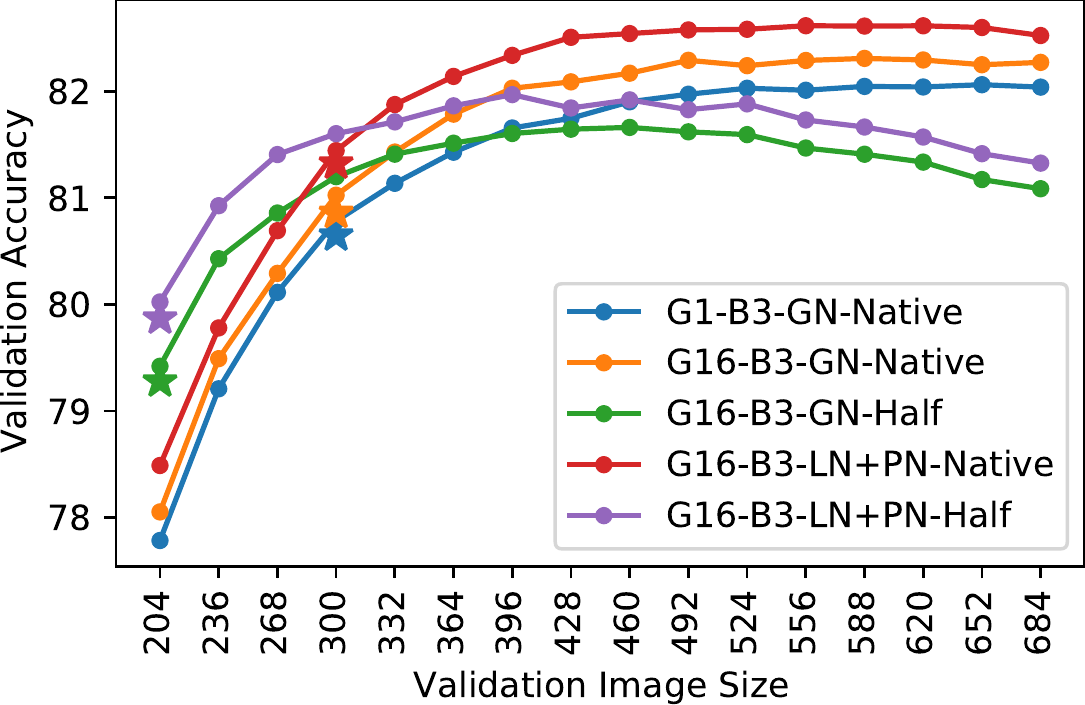}}%
\caption{EfficientNet-B3 validation accuracy after fine-tuning for a range of fine-tuning/testing image sizes. (a)-(b) Effect of the number of fine-tuned blocks; (c) Results spanning different model and training configurations after fine-tuning with the optimal number of blocks. Label format {G-M-R} with G: group size, M: normalization method, R: training resolution.}%
\label{fig:Train_Test}%
\end{figure*}

\begin{table*}[ht]
\caption{Validation accuracy $\pm$ standard deviation (\%) for different model sizes and training configurations, fine-tuned and tested at the \textbf{native image resolution}. Headers format defined as {G-M-R} for G: group size, M: normalization method, R: training resolution.}
\label{tab:Train_Test_results_N}
\vskip 0.15in
\begin{center}
\begin{small}
\begin{sc}
\begin{tabular}{lrrrrr}
\toprule
Size &          G1-GN-Native &         G16-GN-Native &           G16-GN-Half &      G16-LN+PN-Native &                 G16-LN+PN-Half \\
\midrule
  B0 &  76.4{\tiny $\pm$0.1} &  76.4{\tiny $\pm$0.1} &  76.7{\tiny $\pm$0.2} &  77.0{\tiny $\pm$0.2} &  \textbf{77.3{\tiny $\pm$0.1}} \\
  B1 &  78.4{\tiny $\pm$0.0} &  78.6{\tiny $\pm$0.1} &  78.9{\tiny $\pm$0.0} &  79.2{\tiny $\pm$0.1} &  \textbf{79.4{\tiny $\pm$0.1}} \\
  B2 &  79.4{\tiny $\pm$0.2} &  79.7{\tiny $\pm$0.1} &  79.8{\tiny $\pm$0.1} &  80.0{\tiny $\pm$0.2} &  \textbf{80.4{\tiny $\pm$0.1}} \\
  B3 &  80.8{\tiny $\pm$0.1} &  81.0{\tiny $\pm$0.1} &  81.2{\tiny $\pm$0.1} &  81.4{\tiny $\pm$0.2} &  \textbf{81.6{\tiny $\pm$0.1}} \\
  B4 &  82.1{\tiny $\pm$0.1} &  82.4{\tiny $\pm$0.1} &  82.6{\tiny $\pm$0.1} &  82.7{\tiny $\pm$0.1} &  \textbf{82.8{\tiny $\pm$0.1}} \\
  B5 &  82.9{\tiny $\pm$0.1} &  83.5{\tiny $\pm$0.1} &  83.5{\tiny $\pm$0.1} &  83.5{\tiny $\pm$0.1} &  \textbf{83.8{\tiny $\pm$0.1}} \\
\bottomrule
\end{tabular}
\end{sc}
\end{small}
\end{center}
\vskip -0.1in

\caption{Validation accuracy $\pm$ standard deviation (\%) for different model sizes and training configurations, fine-tuned and tested at the \textbf{best image resolution}. Best resolution shown in $[\cdot]$. Headers as described above. }
\label{tab:Train_Test_results_B}
\vskip 0.15in
\begin{center}
\begin{small}
\begin{sc}
\begin{tabular}{lrrrrr}
\toprule
Size &                G1-GN-Native &               G16-GN-Native &                 G16-GN-Half &                     G16-LN+PN-Native &              G16-LN+PN-Half \\
\midrule
  B0 &  79.1{\tiny $\pm$0.2} [480] &  79.1{\tiny $\pm$0.1} [416] &  77.8{\tiny $\pm$0.1} [352] &  \textbf{79.4{\tiny $\pm$0.0} [448]} &  78.3{\tiny $\pm$0.1} [384] \\
  B1 &  80.8{\tiny $\pm$0.2} [528] &  80.8{\tiny $\pm$0.0} [528] &  80.0{\tiny $\pm$0.1} [400] &  \textbf{81.3{\tiny $\pm$0.1} [528]} &  80.5{\tiny $\pm$0.1} [400] \\
  B2 &  81.4{\tiny $\pm$0.1} [548] &  81.5{\tiny $\pm$0.1} [516] &  80.8{\tiny $\pm$0.1} [388] &  \textbf{81.8{\tiny $\pm$0.1} [516]} &  81.2{\tiny $\pm$0.0} [420] \\
  B3 &  82.1{\tiny $\pm$0.0} [652] &  82.3{\tiny $\pm$0.1} [588] &  81.7{\tiny $\pm$0.0} [460] &  \textbf{82.6{\tiny $\pm$0.0} [556]} &  82.0{\tiny $\pm$0.1} [396] \\
  B4 &  82.9{\tiny $\pm$0.1} [668] &  83.2{\tiny $\pm$0.1} [572] &  82.7{\tiny $\pm$0.1} [508] &  \textbf{83.4{\tiny $\pm$0.1} [604]} &  83.0{\tiny $\pm$0.1} [444] \\
  B5 &  83.4{\tiny $\pm$0.1} [680] &  84.0{\tiny $\pm$0.1} [648] &  83.6{\tiny $\pm$0.1} [424] &  \textbf{84.0{\tiny $\pm$0.1} [616]} &  83.8{\tiny $\pm$0.0} [488] \\
\bottomrule
\end{tabular}

\end{sc}
\end{small}
\end{center}
\vskip -0.1in
\end{table*}

Figure~\ref{fig:Train_Test} shows the effects of fine-tuning and testing across the full sweep of validation image sizes for EfficientNet-B3. Fine-tuning is shown to improve all networks' accuracies. However, as we increase the number of fine-tuned layers, improvements in validation accuracy mainly occur at large image resolution. This supports the intuition that the further we get from the training distribution, the more layers we need to fine-tune. Although the best validation accuracies are seen at high resolution, it may be practical to accept a slight degradation of performance in return for the reduced cost of inference at lower resolutions.

Interestingly, we see different fine-tuning characteristics for the G1- and G16-EfficientNet models in Figure~\ref{fig:Train_Test_a}. With little or no fine-tuning, the G16 model performs significantly better at large image resolutions. While this performance gap is largely eliminated as more layers get fine-tuned, it may again suggest that the G16 model has beneficial generalisation properties compared to the vanilla G1 model. In Figure~\ref{fig:Train_Test_c}, we also see that the benefits provided by PN are robust across image sizes.

When comparing the different training resolutions in Figure~\ref{fig:Train_Test_b} and Table~\ref{tab:Train_Test_results_N}, we surprisingly find that \textbf{models trained at half image size outperform models trained at native image size} when both are tested at native image size. We initially hypothesised that this could arise from the mismatch in the distribution of object sizes between the standard preprocessing of ImageNet at training and validation~\cite{Touvron19}. However, this difference was persistent even after fine-tuning the native-size training with validation preprocessing, suggesting that low-resolution training may itself offer a beneficial regularisation effect.

Comparing the different model sizes in Table~\ref{tab:Train_Test_results_B}, we see that, when testing at the \textit{best} resolution, the highest accuracy is achieved by training at the larger, native image size. This, however, requires larger resolutions to be used in testing, which would adversely affect the inference cost.

This strongly suggests that we should \textit{always} be fine-tuning and testing on larger images than trained on to achieve the best accuracy.
Conversely, this also suggests that, to achieve maximum performance at a target test resolution, the model should be trained with smaller images. Half the size appears to be a reasonable choice, though a full sweep of training resolutions would be required to investigate whether this is indeed optimal. On the other hand, fine-tuning on lower resolutions than trained on does not seem to offer any clear benefits.

\subsection{Training Efficiency}

\begin{figure*}[p!]%
\centering
\subfloat[a][]{\label{fig:Training_efficiency_Na}\includegraphics[width=0.3\linewidth]{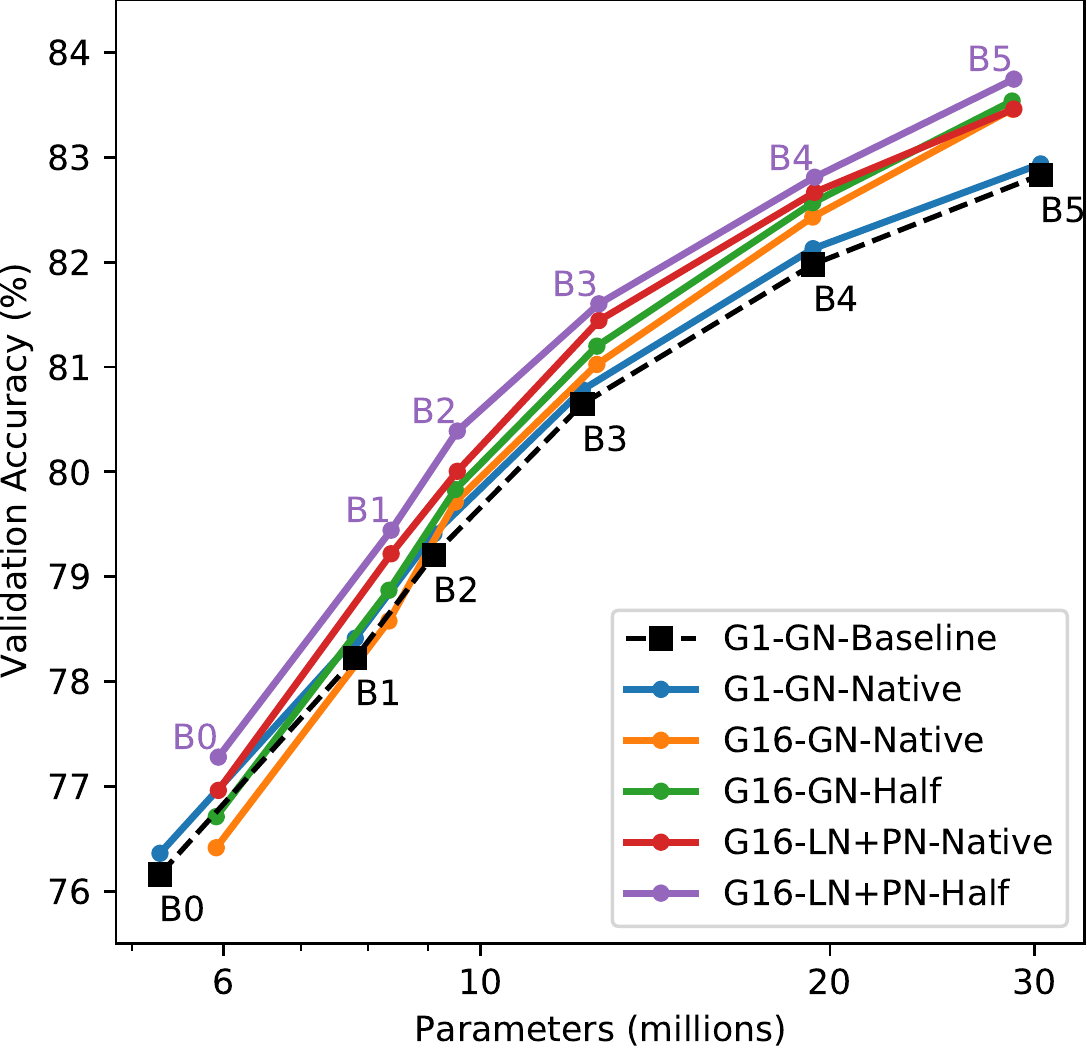}}\quad
\subfloat[b][]{\label{fig:Training_efficiency_Nb}\includegraphics[width=0.3\linewidth]{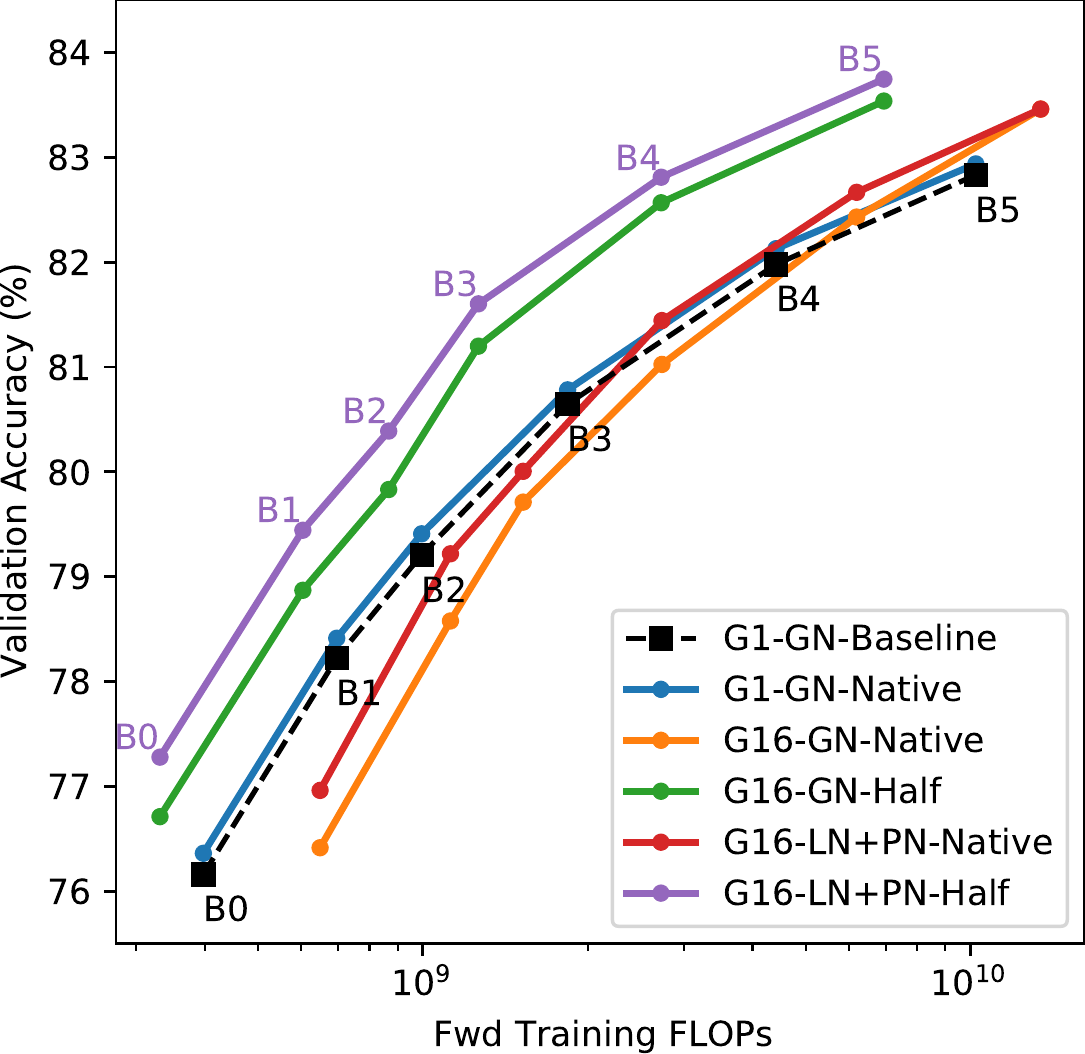}}\quad
\subfloat[c][]{\label{fig:Training_efficiency_Nc}\includegraphics[width=0.3\linewidth]{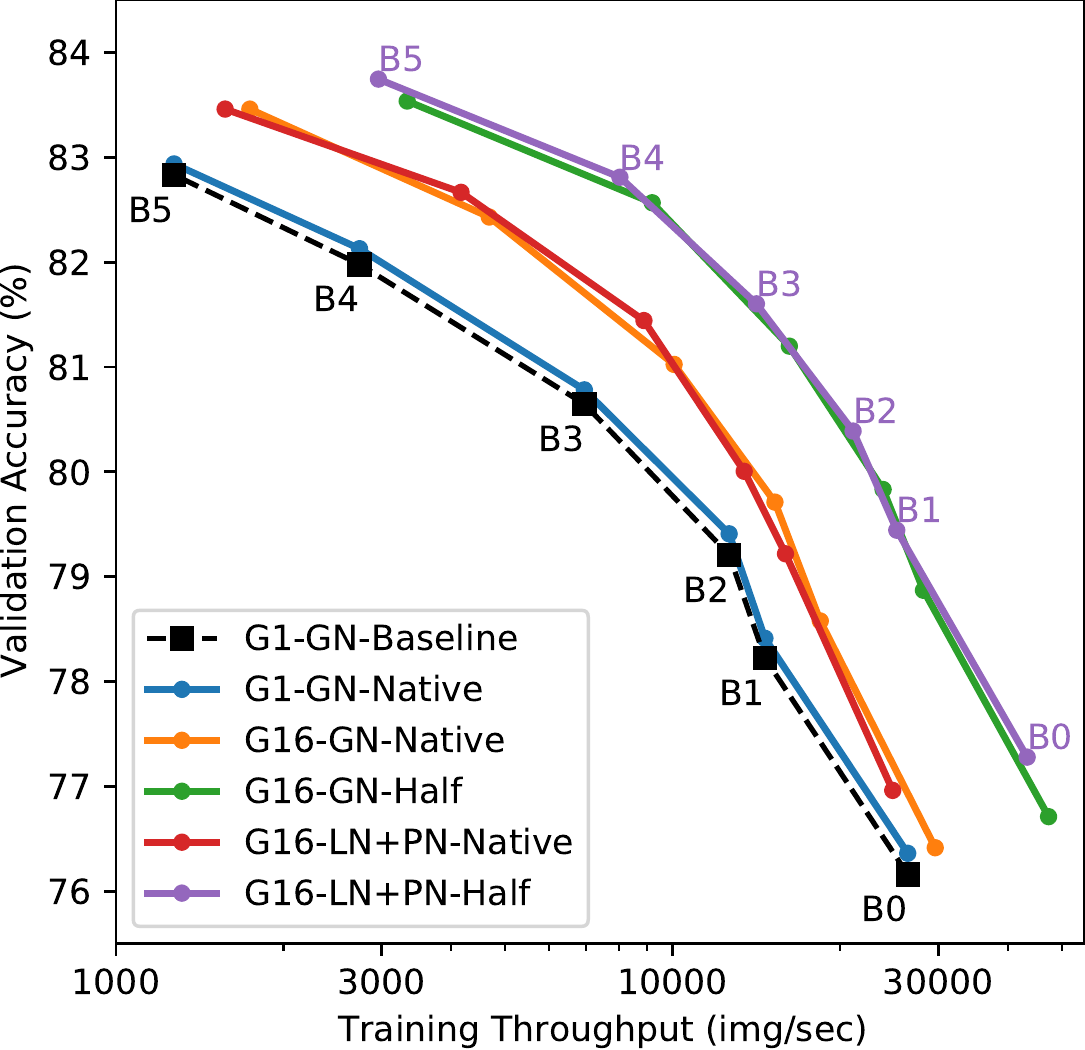}}\quad
\caption{Comparison of parameter (a), theoretical (b) and practical (c) efficiency when fine-tuning and testing at the \textbf{native image resolution}. Label format defined as {G-M-R} for G: group size, M: normalization method, R: training resolution. G1-GN-Baseline case uses native resolution but is not fine-tuned.}%
\label{fig:Training_efficiency_N}%
\end{figure*}
\begin{figure*}[p!]%
\centering
\subfloat[a][]{\label{fig:Training_efficiency_Ba}\includegraphics[width=0.3\linewidth]{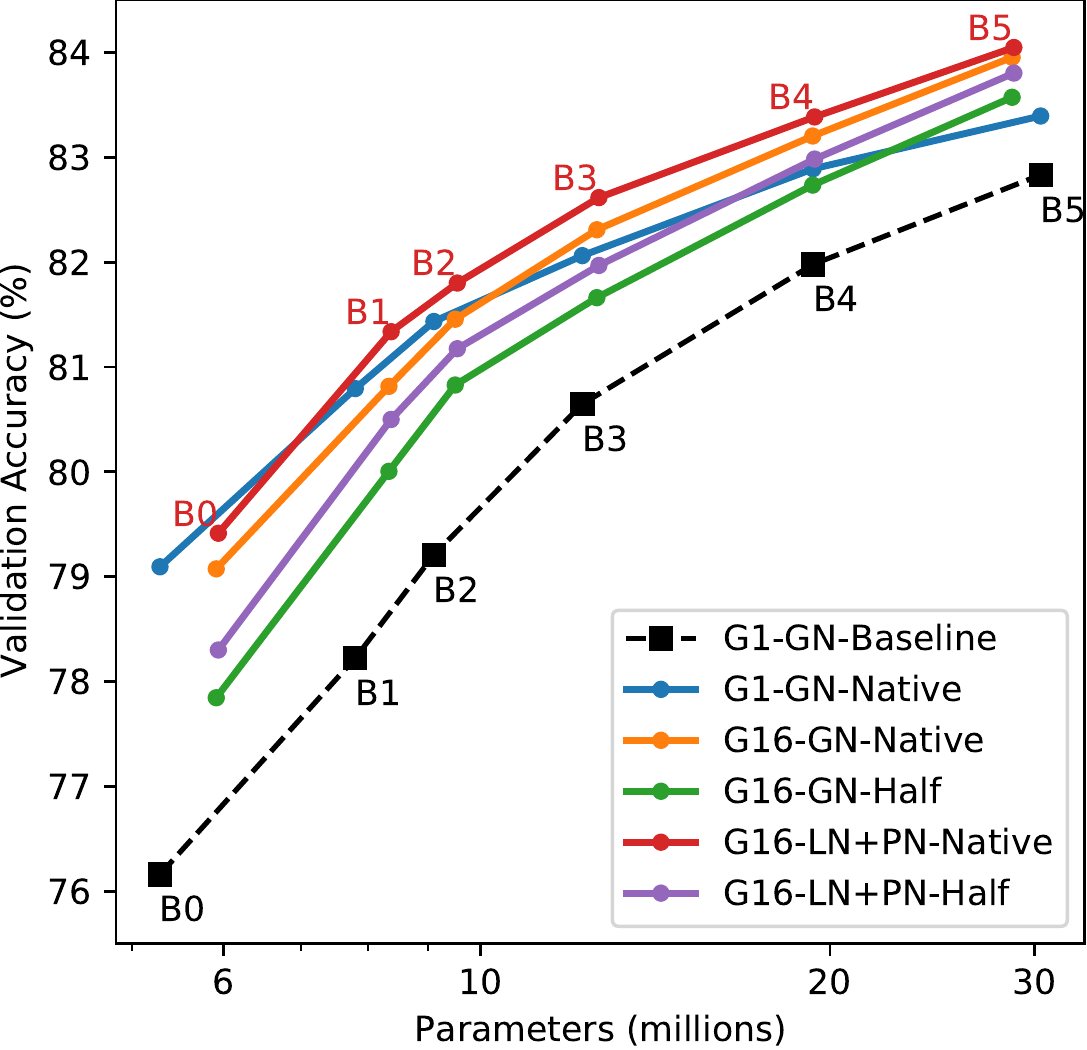}}\quad
\subfloat[b][]{\label{fig:Training_efficiency_Bb}\includegraphics[width=0.3\linewidth]{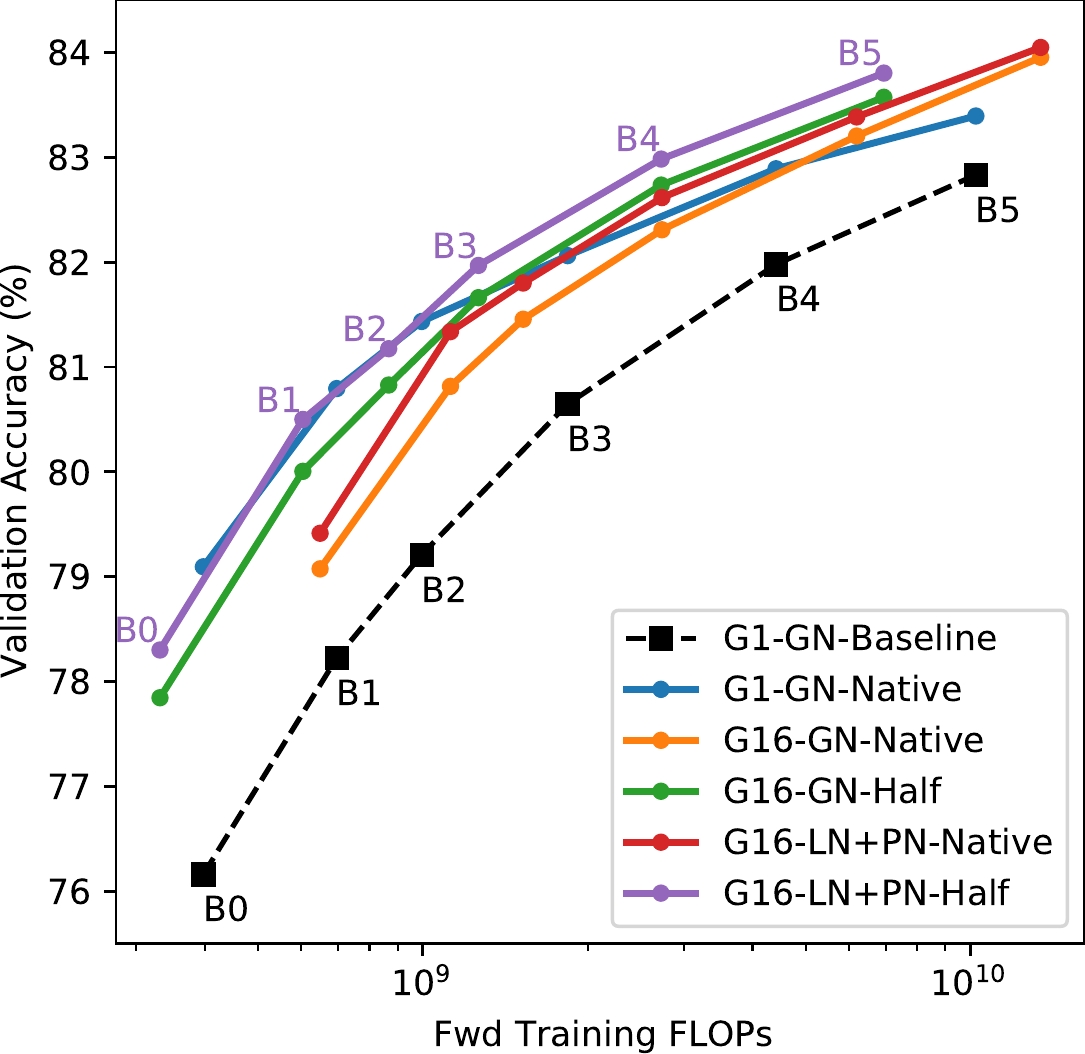}}\quad
\subfloat[c][]{\label{fig:Training_efficiency_Bc}\includegraphics[width=0.3\linewidth]{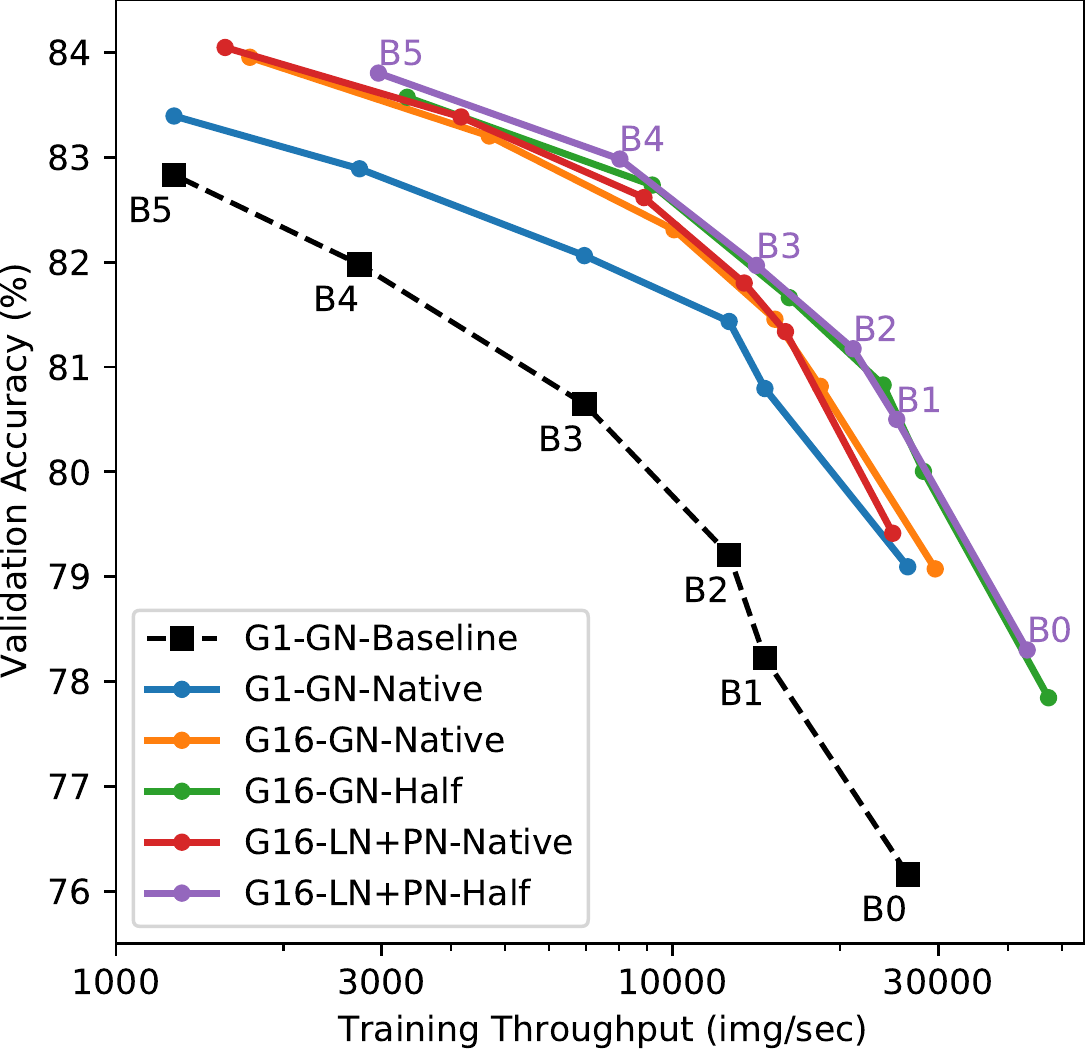}}\quad
\caption{Comparison of parameter (a), theoretical (b) and practical (c) efficiency when fine-tuning and testing at the \textbf{best image resolution}. Label format same as Figure~\ref{fig:Training_efficiency_N}.}%
\label{fig:Training_efficiency_B}%
\end{figure*}
\begin{figure*}[p!]
\centering
\subfloat[a][]{\label{fig:Inference_efficiency_a}\includegraphics[width=0.3\linewidth]{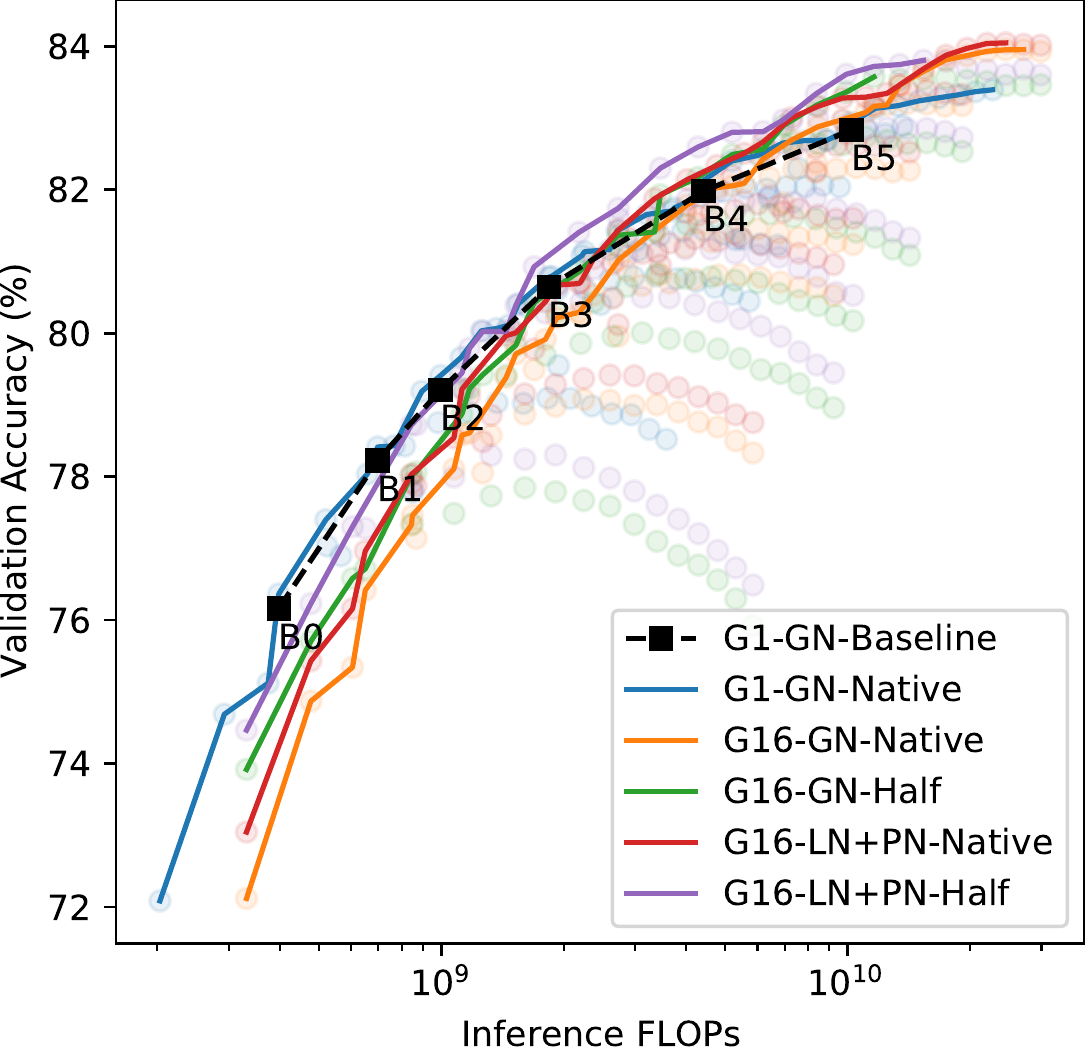}}\quad
\subfloat[b][]{\label{fig:Inference_efficiency_b}\includegraphics[width=0.3\linewidth]{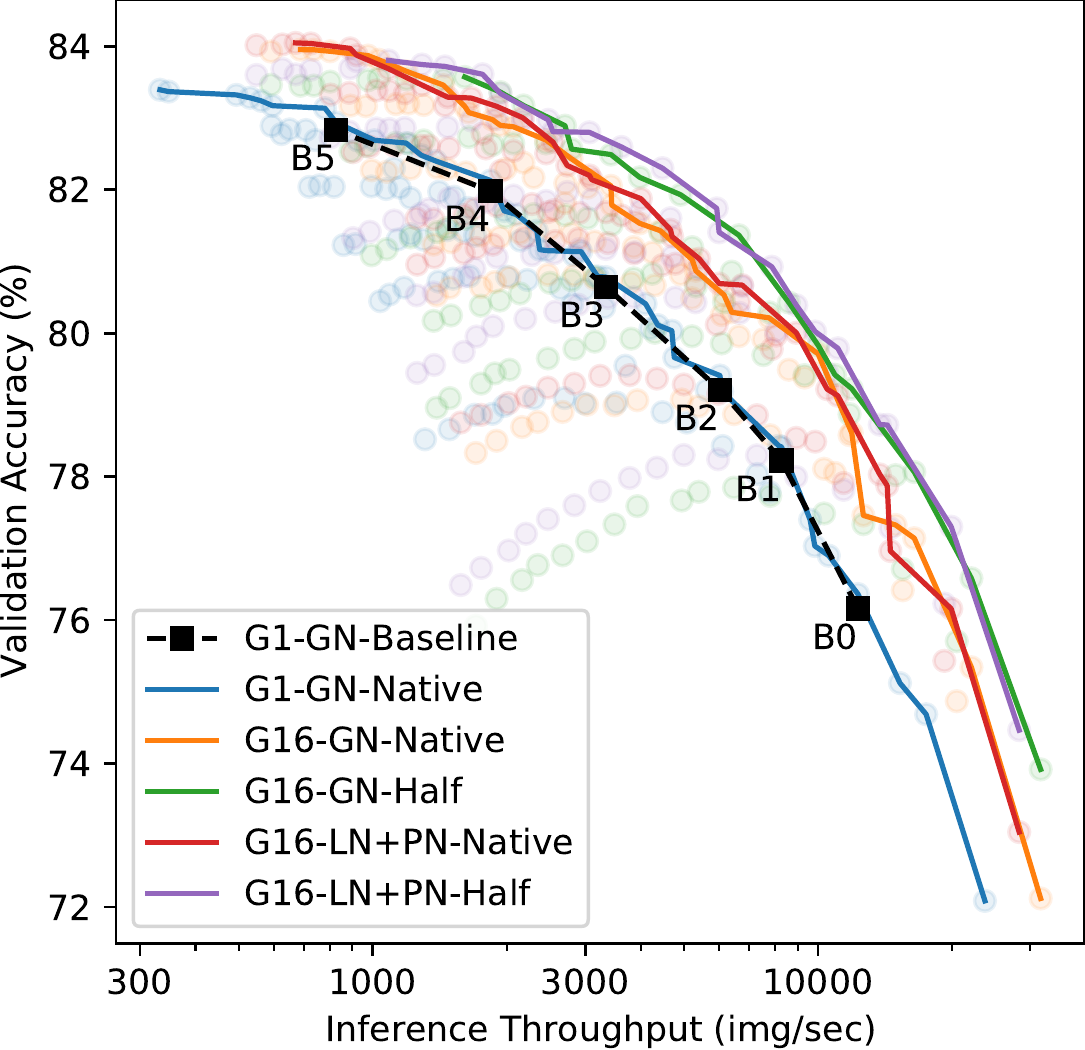}}\quad
\subfloat[c][]{\label{fig:Inference_efficiency_c}\includegraphics[width=0.3\linewidth]{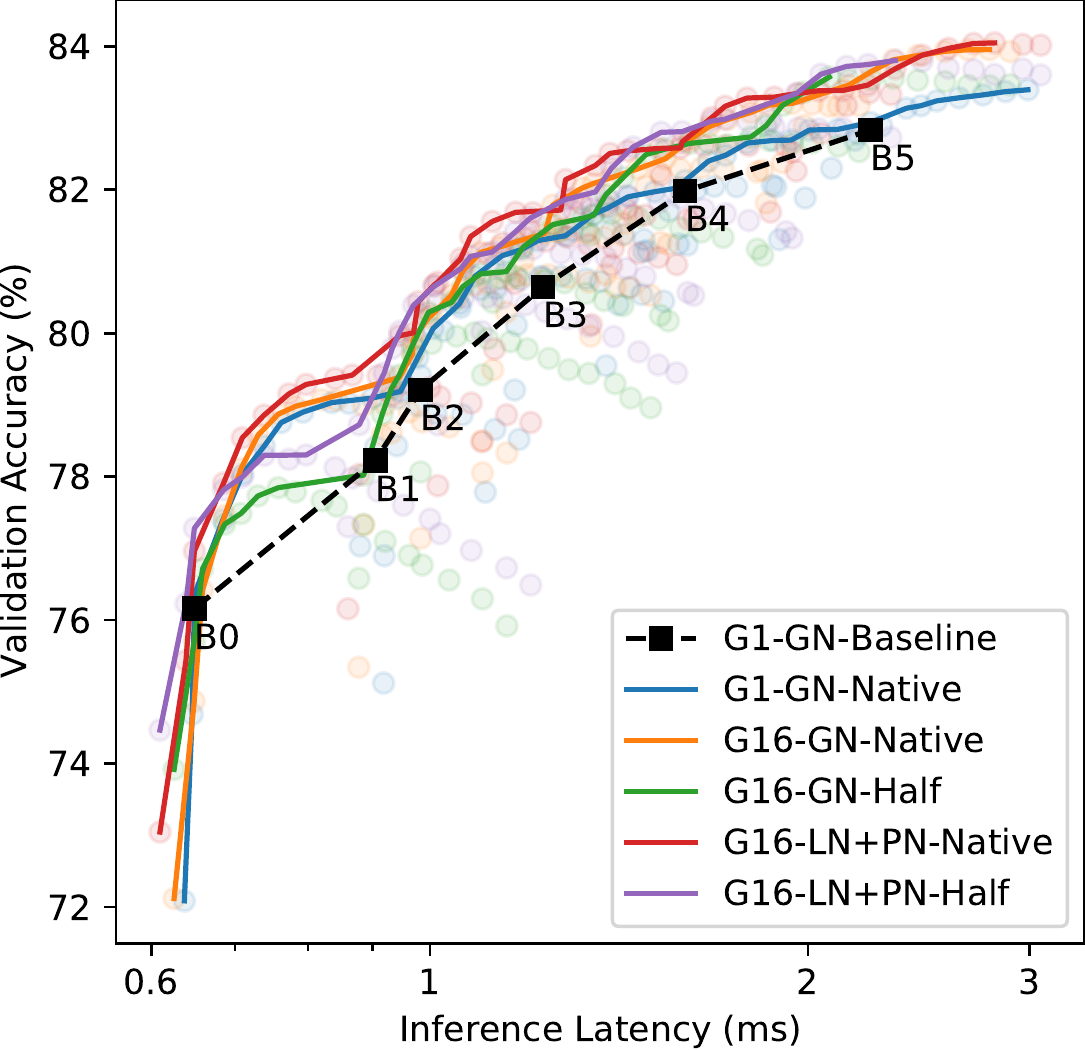}}\quad
\caption{Comparison of theoretical inference efficiency (a) and two measures of practical inference efficiency: throughput at maximum batch size (b) and latency at batch size 1 (c). Points span different model sizes and different fine-tuning/testing resolutions; lines are Pareto fronts. Label format same as Figure~\ref{fig:Training_efficiency_N}.}
\label{fig:Inference_efficiency}
\end{figure*}

Figures~\ref{fig:Training_efficiency_N} and \ref{fig:Training_efficiency_B} present the results of Tables~\ref{tab:Train_Test_results_N} and \ref{tab:Train_Test_results_B} from the perspective of efficiency. Primarily, we see that there are clear efficiency benefits to training at half resolution. In the context of native resolution testing (Figure~\ref{fig:Training_efficiency_N} and Table~\ref{tab:Train_Test_results_N}), we see that the $2\times$ improvement in theoretical efficiency is directly reflected in terms of practical throughput. When testing with the \textit{best} image resolution (Figure~\ref{fig:Training_efficiency_B} and Table~\ref{tab:Train_Test_results_B}), the gap between the training resolutions is decreased significantly, suggesting that the increased ratio of test to training resolution is a significant factor.

For the G1-EfficientNet case, we see a much worse translation of theoretical efficiency to practical gains compared to G16-EfficientNet. This strongly supports the choice of adopting group convolutions.

For the LN+PN configurations, we see theoretical efficiency benefits due to the increased accuracy over the corresponding models with GN. In practice, however, we find that the efficiency curves are not substantially improved due to approximately a 10\% throughput cost. Importantly we do see a benefit in the number of parameters used in these cases and hope that the theoretical efficiency gains can be realised in practice with further software optimisation.

For the most efficient LN+PN-Half configuration, tested at the best image resolution, we see a $3\times$ to $7\times$ increase in training throughput compared to the non-fine-tuned baseline.

\subsection{Inference Efficiency}

Figure~\ref{fig:Inference_efficiency} presents efficiency results for inference, considering the maximum batch size throughput and batch size 1 for latency measurements. 
Here we see that half resolution training produces more theoretically efficient models for inference, due to better accuracy at smaller image size (see Figure~\ref{fig:Train_Test}). We also find that these Pareto optimal \textit{Half} configurations still outperform the \textit{Native} configurations for training throughput, highlighting that efficiency benefits can be achieved simultaneously in both training and inference with a single test resolution.

While this translates well to practical throughput efficiency at maximum batch size, for practical latency efficiency at batch size 1, the native training resolution cases fare better. In this case, larger image sizes are less strongly penalised, since increased memory cost does not result in the use of a lower batch size.

The G1-EfficientNet variant again does not manage to convert the theoretical efficiency gains into practical throughput gains as well as the G16 variant. G1 does, however, perform comparably for the minimum latency case. Similarly to the large image size case, this suggests that this variant benefits from lower penalisation of large memory costs.

Across both practical cases, the G16-LN+PN models improve efficiency over the baseline models, achieving a $1.9\times$ to $3.6\times$ throughput increase and up to $1.4\times$ latency improvement.  

\section{Conclusion}

We have investigated three distinct, complementary techniques for improving the practical efficiency of EfficientNet models. Based on a broad set of experimental results, each of these techniques has been shown to provide improved validation accuracy for the same computational cost
and/or equivalent accuracy at reduced training cost. When using all three methods in combination, we achieved throughput benefits of up 7$\times$ and 3.6$\times$ compared to the baseline at the same validation accuracy.

\section{Concurrent Work}
Some concurrent work has drawn similar conclusions to those we present here.

\citet{Tan21} also investigated how to speed up EfficientNet in practice. In agreement with this work they found that training on lower resolution images provided significant gains in practical efficiency. However they chose to progressively increase the resolution throughout training rather than finetune it at the end. They also modified the model itself, and in similar fashion to this work, chose to \say{densify} the MBconv blocks. They did this by replacing the depthwise spatial convolution and the second pointwise convolution with a single dense, spatial convolution. They also found benefits from reducing the expansion factor to 4.

\citet{Brock21a} looked at scaling up a ResNet-like model with batch independent normalization. They proposed a new approach to normalization based on weight standardization~\cite{Qiao19} and layerwise scaling. However, based on \citet{Brock21,Labatie21} we believe this approach would not be effective on EfficientNet. They also highlighted that training on smaller images than the original EfficientNet work was key to achieving good training efficiency. \citet{Bello21} also showed strong evidence to support the benefits of using lower resolution images in training.

\bibliography{ipu_efficientnet}
\bibliographystyle{icml2020}

\appendix
\section{Arithmetic Intensity}
\label{sec:arithemtic_intensity}

One simple but useful metric when considering the practical performance of neural networks is the ratio of the compute (FLOPs) to memory transfer (bytes) required for the model. This dictates the memory bandwidth requirements of a hardware platform in order to ensure that the arithmetic compute units are sufficiently utilised. 

Given the assumption that weight and activation state must always be transferred, we can compute the approximate intensity of an arbitrary convolution with batch size $B$, kernel size $k\times k$, field size $f\times f$, group size $G$, number of groups $N$ and stride $s$ as
\begin{align}
    I &= \frac{\textrm{FLOPs}}{\textrm{Weight Mem}+\textrm{Act Mem}} \\
    &= \frac{G^2k^2Bf^2N/s}{k^2G^2N+Bf^2GN} \\
    &= \frac{Gk^2Bf^2}{s(Gk^2+Bf^2)}
\end{align}

From this we can see that arithmetic intensity has a positive monotonic relationship with $G$, $k$, $b$ and $f$ and a negative monotonic relationship with $s$. Interestingly the arithmetic intensity does not depend on the number of groups $N$.

\section{TPU Experiments}
\label{sec:TPU Experiments}

Machine learning experiments are often underspecified in the corresponding papers. To compare our models in the closest possible setting to the original \say{vanilla} EfficientNets, we used the public EfficientNet repository to run experiments using Google Cloud TPUs.\footnote{\url{https://github.com/tensorflow/tpu/tree/master/models/official/efficientnet}, accessed 2020-13-12.} For these experiments, we ran EfficientNet experiments for B0 and B2, using the settings recommended in the repository. We also implemented and ran our G16 version of the network in this codebase. We ran each experiment (B0/B2; G1/G16) with and without \say{Autoaugment}, the recommended augmentation scheme. 

\begin{table}[htbp]
\caption{Validation accuracy $\pm$ standard deviation (\%) for batch-normalized results on the TPU. For the \say{AutoAugment} cases, the AutoAugment hyperparameters were as specified in the code repository. }
\label{tab:tpu}
\vskip 0.15in
\centering
\begin{tabular}{@{}llll@{}}
\toprule
            & Size & G1        & G16            \\ \midrule
Baseline    & B0   & 76.9{\tiny$\pm$0.06} & 76.8{\tiny$\pm$0.13} \\
AutoAugment & B0   & 77.2{\tiny$\pm$0.09} & 76.7{\tiny$\pm$0.15} \\
Baseline    & B2   & 79.4{\tiny$\pm$0.04} & 79.5{\tiny$\pm$0.06} \\
AutoAugment & B2   & 80.0{\tiny$\pm$0.03} & 79.7{\tiny$\pm$0.12} \\ \bottomrule
\end{tabular}
\end{table}

The results of these experiments are presented in Table~\ref{tab:tpu}. For the cases without AutoAugment, the final performance of the \say{vanilla} networks is indistinguishable from the G16 versions. However, enabling AutoAugment appears to favour the \say{vanilla} networks. AutoAugment in general appears to be finely tuned -- for instance, the AutoAugment parameters published in the public repository do not match the values published in~\citet{Cubuk18}, suggesting that it has been re-tuned for G1 (vanilla) EfficientNets. Therefore, we would expect a similar tuning effort to yield augmentation parameters that would produce comparable results for our G16 networks.

\end{document}